\documentclass[10pt,twocolumn,letterpaper]{article}

\usepackage{wacv}              

\usepackage{graphicx}
\usepackage{amsmath}
\usepackage{amssymb}
\usepackage{booktabs}
\usepackage{colortbl}
\usepackage{tabularx}
\usepackage{array}
\usepackage{multirow}
\usepackage[dvipsnames]{xcolor}
\usepackage{caption}
\usepackage[accsupp]{axessibility}

\usepackage[pagebackref,breaklinks,colorlinks]{hyperref}

\usepackage[capitalize]{cleveref}
\crefname{section}{Section}{Sections}
\Crefname{section}{Section}{Sections}
\Crefname{table}{Table}{Tables}
\crefname{table}{Table}{Tables}
\Crefname{figure}{Figure}{Figures}
\crefname{figure}{Figure}{Figures}

\DeclareMathOperator*{\argmax}{arg\,max}

\def\wacvPaperID{1908} 
\def\confName{WACV}
\def\confYear{2024}
\begin{document}

\title{Instruct Me More! Random Prompting for Visual In-Context Learning}


\author{Jiahao Zhang$^{1}$, Bowen Wang$^{2}\thanks{Corresponding author.}$, Liangzhi Li$^{2}$, Yuta Nakashima$^{2}$, Hajime Nagahara$^{2}$\\
Osaka University, Japan\\
{\tt\small $^1$jiahao@is.ids.osaka-u.ac.jp}\\
{\tt\small $^2$\{wang, li, n-yuta, nagahara\}@ids.osaka-u.ac.jp}
}
\maketitle

\begin{abstract}

Large-scale models trained on extensive datasets, have emerged as the preferred approach due to their high generalizability across various tasks. In-context learning (ICL), a popular strategy in natural language processing, uses such models for different tasks by providing instructive prompts but without updating model parameters. This idea is now being explored in computer vision, where an input-output image pair (called an in-context pair) is supplied to the model with a query image as a prompt to exemplify the desired output. The efficacy of visual ICL often depends on the quality of the prompts. We thus introduce a method coined \textbf{\underline{In}}struct \textbf{\underline{Me}} \textbf{\underline{Mo}}re (InMeMo), which augments in-context pairs with a learnable perturbation (prompt), to explore its potential. Our experiments on mainstream tasks reveal that InMeMo surpasses the current state-of-the-art performance. Specifically, compared to the baseline without learnable prompt, InMeMo boosts mIoU scores by 7.35 and 15.13 for foreground segmentation and single object detection tasks, respectively. Our findings suggest that InMeMo offers a versatile and efficient way to enhance the performance of visual ICL with lightweight training. Code is available at \url{https://github.com/Jackieam/InMeMo}.

\end{abstract}

\section{Introduction}
\label{sec:intro}

The advancement of large-scale models has been profound in recent years. They have demonstrated remarkable abilities to generalize and hold potential for diverse downstream tasks \cite{largemodel, dosovitskiy2020image, clip}. Models such as ChatGPT/GPT-3 \cite{gpt3}, have emphasized the intrinsic capacity of \textit{in-context learning} (ICL) for Natural Language Processing (NLP) tasks \cite{rubin2021learning, gonen2022demystifying, wu2022self, sorensen2022information, honovich2022instruction, wang2022self, min2021noisy}. ICL allows models to undertake new tasks using prompts to predict unseen samples, eliminating the need for model parameter adjustments and reducing training costs. While teeming with potential as a fundamental approach for real-world applications of large-scale models, ICL for computer vision tasks still remains in its exploratory stages \cite{supicl}.

\begin{figure}[t]
  \centering
   \includegraphics[width=1\linewidth]{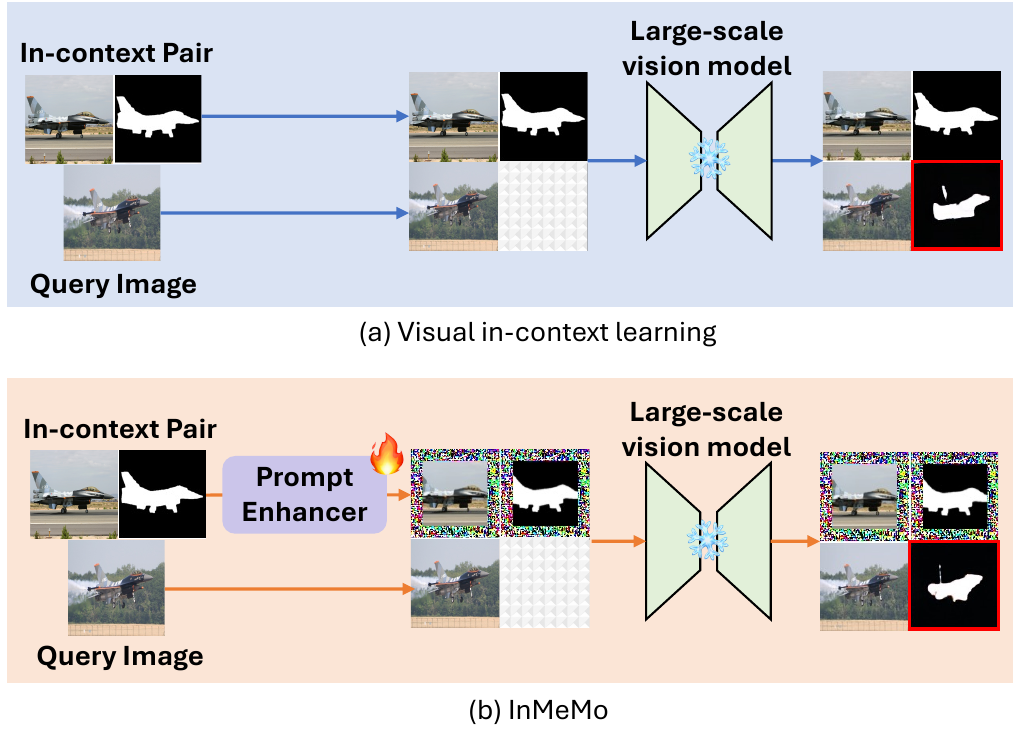}
   \caption{A schematic comparison of current visual ICL and InMeMo. (a) \textcolor{RoyalBlue}{Visual ICL} compiles a query image and in-context pair to create a four-cell grid canvas with an empty cell for a prediction (located in the bottom-right cell in this diagram), which forms a \textit{prompt} for visual ICL. The prediction (depicted in the red box) is obtained by feeding the prompt into a frozen large-scale vision model. (b) \textcolor{Orange}{InMeMo} additionally uses a \textit{learnable prompt}, which is a perturbation to amend the distribution of prompts.}
   \label{fig:icl_concept}
\end{figure}

\begin{figure}[t]
  \centering
   \includegraphics[width=1\linewidth]{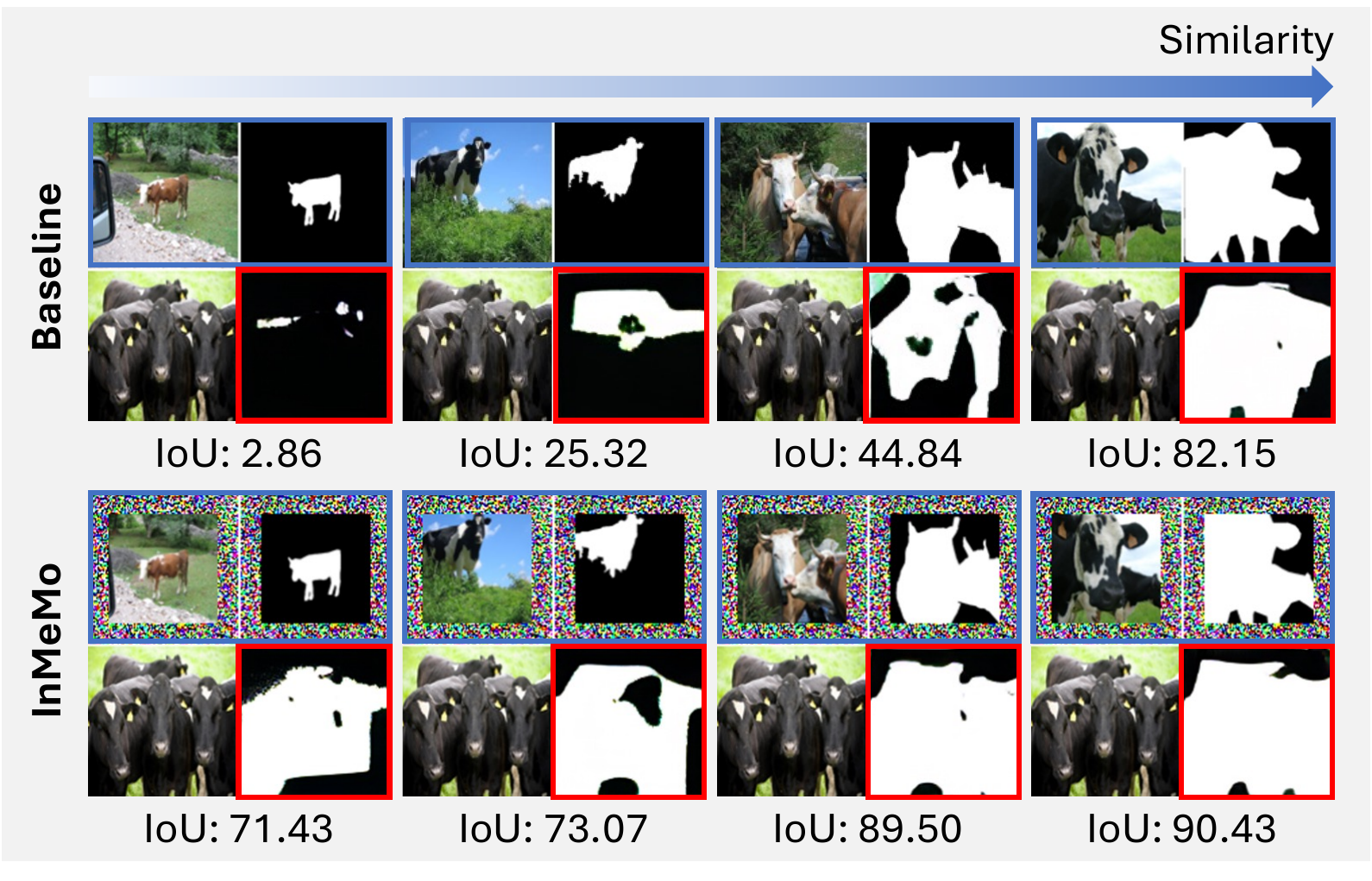}
   \caption{The performance of visual ICL on a foreground segmentation task. \textcolor{RoyalBlue}{Blue boxes} and \textcolor{red}{red boxes} are in-context pairs and predicted label images (query images are not marked). The in-context pair largely affects the performance. Without a learnable prompt, the performance depends much on the similarity of the query and in-context images. InMeMo, which uses learnable prompts, generates more consistent predictions.}
   \label{fig:prompt_selection}
\end{figure}

MAE-VQGAN \cite{maevqgan} marks a pioneering effort, showcasing the feasibility of ICL in computer vision across various tasks, such as image segmentation, inpainting, and style transfer. This method employs visual prompts in a grid format as in \cref{fig:icl_concept}(a), comprising a query image and an input-output pair, called an in-context pair, that exemplifies the task to be solved with an input image and its corresponding label image. Some studies emphasize the pivotal role of in-context pairs for better instructing a model in generating desired outputs. That is, visual ICL demands an in-context image that is similar to the query image in terms of its semantics, viewpoint, \etc \cite{supicl}) as shown in \cref{fig:prompt_selection}, making in-context pair retrieval an indispensable step.

Despite the notable success \cite{supicl, promptself} achieved, retrieved in-context pairs may not be optimal due to the finite size of the dataset to retrieve and a gap between prompts and knowledge in a large-scale vision model. This observation inspires us with an idea: \textit{Can we transform the prompt to better instruct the model for downstream tasks in visual ICL?}

Learnable prompting\footnote{In \cite{firstvp}, the idea of adding a learnable pixel-level perturbation to images is called \textit{visual prompting}; however, as our work also involves visual prompts consisting of an in-context pair and a query, we rephrase a learnable perturbation with a \textit{learnable prompt}.} \cite{firstvp,ilmvp, blackvip}, which applies a transformation to the model's inputs without modifying the model itself for adapting to various downstream tasks, shows superior performance in image classification. This method, which can be seen as a type of parameter-efficient transfer learning (PETL) \cite{vpt, lester2021power, prompttuning2, coop}, is particularly effective in large-scale models compared to fine-tuning, primarily because large-scale models involve enormous training parameters and require significant computational resources even for fine-tuning \cite{ilmvp, firstvp, elsayed2018adversarial, neekhara2022cross}. Notably, the learnable prompt has demonstrated a robust capability to fit data, even when there are significant discrepancies present \cite{blackvip, ilmvp, firstvp}.

We are thus pioneering our visual ICL method \textbf{\underline {In}}struct \textbf{\underline{Me}} \textbf{\underline{Mo}}re (\textbf{InMeMo}) for instructing a large-scale model by a visual learnable prompt. After in-context pair retrieval, we amend the pair with our prompt enhancer, as in \cite{firstvp}. As with the existing visual ICL methods, InMeMo compiles the enhanced pair and the query image into a single image called canvas, which is then fed into a pre-trained large-scale vision model \cite{maevqgan}. Our learnable prompt is trained in a supervised manner to generate the corresponding ground-truth label image for the query.

\textbf{Contributions.} InMeMo is a PETL approach, enjoying a lightweight training process. A learnable prompt dedicated to a given downstream task translates the distribution of entire prompts to make them more task-specific and improve the large-scale model's encoding and decoding efficiency. Our experimental results successfully support our claim by showing new state-of-the-art (SOTA) performance in foreground segmentation and single object detection tasks. Although training is indispensable for InMeMo, it effectively alleviates the challenges posed by lower-quality visual prompts.

\begin{figure*}[t]
  \centering
   \includegraphics[width=1\linewidth]{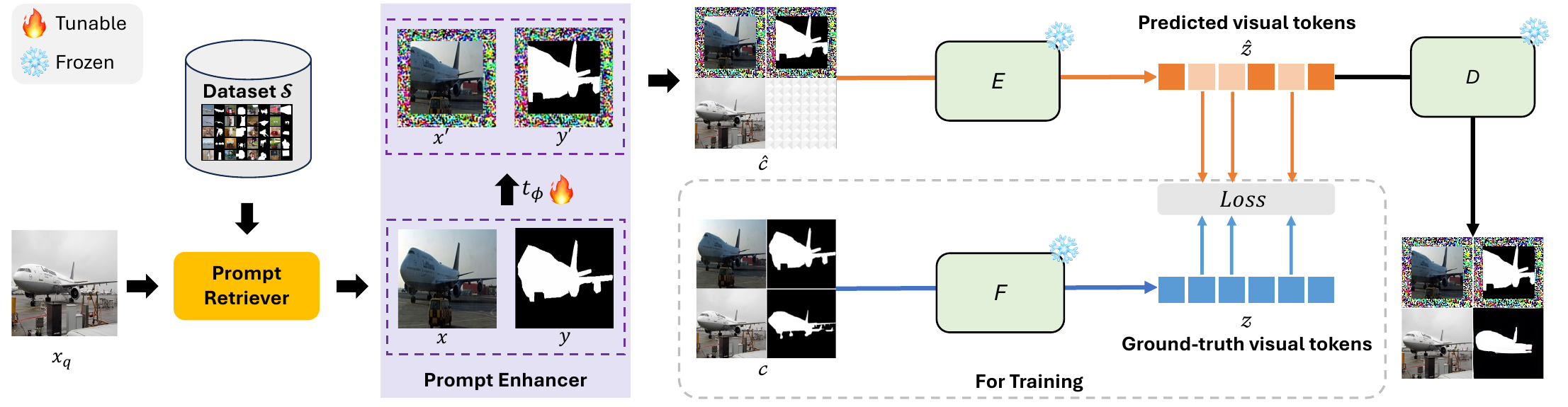}
   \caption{The overall framework of the proposed InMeMo method. First, we employ the Prompt Retriever from the dataset $\mathcal{S}$ to select an in-context pair $(x, y)$ for a query image $x_\text{q}$. We then use a \textcolor{Plum}{Prompt Enhancer $t_\phi(\cdot)$} to add perturbations to the in-context pair separately to obtain an enhanced in-context pair $(x', y')$. We create a four-cell grid canvas $\hat{c}$ containing $(x', y', x_\text{q}, \varnothing)$, with an empty cell at the bottom right. The $\hat{c}$ is fed into a frozen {\tt MAE-VQGAN} ($E$) to generate predicted visual tokens \textcolor{Orange}{$\hat{z}$} containing the empty cell in $\hat{c}$. For visualized prediction, the \textcolor{Orange}{$\hat{z}$} is decoded to visual pixels by the decoder of VQGAN ($D$). To train our InMeMo, a ground-truth canvas $c$ containing $(x, y, x_\text{q}, y_\text{q})$ is fed into a pre-trained encoder of VQGAN ($F$) to generate ground-truth visual tokens \textcolor{Cerulean}{$z$}. We calculate the cross-entropy loss upon the empty cell to \textbf{only} update the Prompt Enhancer parameter $\phi$.}
   \label{fig:inmemo}
\end{figure*}
\section{Related Work}
\subsection{In-Context Learning}

ICL is a recent paradigm in NLP for large language models (LLMs), like GPT-3 \cite{gpt3}. With several pre-defined input-output pairs for a specific task, this approach enables an autoregressive model to enhance performance without tuning model parameters for inference \cite{promptself}. ICL has been verified to be strong enough with several advantages \cite{dong2022survey}, such as offering an interpretable interface to communicate with LLMs \cite{gpt3, liu2021makes, lu2021fantastically}, being similar to human decision-making processes \cite{winston1980learning}, and instantiating a language model as a service \cite{sun2022black}. It also leads to new applications in various fields \cite{iclnlp1, learn2learn, kim2022self}, such as solving mathematics reasoning problems \cite{mathicl}, question answering \cite{learn2learn, press2022measuring}, and compositional generalization \cite{an2023context, hosseini2022compositional}.

In the field of computer vision, ICL is still a new concept with limited existing work \cite{alayrac2022flamingo, supicl, maevqgan, wang2023images}. The challenge in visual ICL lies in specifying the task that the model solves, whereas ICL for NLP uses textual instruction. Bar \etal \cite{maevqgan} proposed to use an input-output image pair, called an in-context pair, with a query image to exemplify the desired output. This combination of them into one image casts the given task as a specific inpainting task. Subsequently, Zhang \etal \cite{supicl} proposed to train a prompt selection model in a supervised manner and demonstrated prompt (in-context pair) selection, and the number of prompts provided to the model is the key to improving the performance of visual ICL. Sun \etal \cite{promptself} suggested using pixel-level in-context pair retrieval for prompt selection. Additionally, they investigated eight different arrangements of the in-context pair and query image and fused the results to enhance ICL performance.

In-context pairs have been proven essential for optimizing performance in downstream tasks \cite{maevqgan, supicl, promptself}. Nonetheless, prior literature has not yet investigated the transformation of the in-context pair to enhance the performance of visual ICL. We aim to explore the potential benefits of introducing learnable perturbation to in-context pairs for improving downstream task performance.

\subsection{Learnable Prompting}

In NLP, prompting can be used to guide LLMs to better adapt to downstream tasks \cite{liu2023pre}. For instance, GPT-3 \cite{gpt3} has shown outstanding generalization ability for different downstream tasks, but costly manually-designed prompting is often necessary. Furthermore, full fine-tuning demands enormous computational resources due to large model sizes. PETL optimizes a small subset or an additional set of parameters of LLMs to specific downstream tasks as in adapter \cite{pfeiffer2020adapterfusion, houlsby2019parameter} and prompt tuning \cite{lester2021power, hu2021knowledgeable}, to achieve competitive performance compared to full fine-tuning.

Due to the exceptional performance of PETL in the field of NLP, numerous previous studies have endeavored different attempts in vision \cite{vpt, tsai2020transfer, blackvip} and vision-and-language models \cite{clip, coop, cocoop}. This is usually accompanied by partially fine-tuning the model or adding learnable prompts to the input image. As a latter approach, Bahng \etal \cite{firstvp} proposed incorporating learnable pixel-level input-independent visual prompting (VP) into the input image to enhance the transferability of large-scale frozen models, such as CLIP \cite{clip}, to downstream tasks. This optimization process only involves a significantly smaller set of parameters than the large-scale model, making VP a well-suited extension for visual ICL. This paper explores the potential of VP for this purpose.

\section{Method}

Let $\mathcal{S}=\{(x, y)\}$ denote a dataset of pairs of an input image $x$ and a label (output) image $y$ for a specific downstream task, where $|S| = n$. Given this dataset and a query image $x_\text{q}$ as input, a prediction $y_\text{q}$ of the task is generated.

\Cref{fig:inmemo} shows an overview of InMeMo. A query image $x_\text{q}$ is fed into the prompt retriever to find an in-context pair $(x, y)$ from dataset $\mathcal{S}$. The prompt enhancer then takes them to obtain a pair $(x', y')$ with a learnable prompt. The pair is concatenated with the query image $x_\text{q}$ to form a four-cell grid canvas, denoted by a quadruple $(x', y', x_\text{q}, \varnothing)$, where $\varnothing$ represents an empty cell. The canvas is fed into a frozen pre-trained large-scale vision model $E$ to obtain visual tokens $\hat{z} = E(x', y', x_\text{q}, \varnothing)$. The visual tokens corresponding to the empty cell encode the prediction $\hat{y}_\text{q}$ of the task. Decoder $D$ gives prediction $\hat{y}_\text{q}$ as $\hat{y}_\text{q} = D(\hat{z})$. 

The key component in InMeMo is the prompt enhancer, denoted by $t_\phi$, with a set $\phi$ of learnable parameters. We train the prompt enhancer with dataset $\mathcal{S}$ so that it is instructive enough to specify the task even when the retriever cannot find an in-context pair with sufficient quality. 

\subsection{Prompt Retriever}

Finding a high-quality in-context pair for a given query image is non-trivial for better performance \cite{supicl}. Our prompt retriever follows pixel-level retrieval in \cite{promptself} for prompt selection. We first use an off-the-shelf feature extractor (\eg, CLIP visual encoder \cite{clip}) to obtain $\ell_2$-normalized visual features of query image $x_\text{q}$ and of in-context image $x \in \mathcal{S}$. The in-context pair in $\mathcal{S}$ whose visual feature is most similar to the query's, is used as in-context pair $(x, y)$, \ie,
\begin{equation}
  (x, y) = \argmax_{(x^\star, y^\star) \in S} v(x_\text{q})^\top v(x^\star),
  \label{eq:prompt_ret}
\end{equation}
where $v(\cdot)$ gives the visual features after the normalization.

\subsection{Prompt Enhancer}

The learnable prompt is conceived in \cite{firstvp}, inspired by the notable successes of prompting in NLP \cite{gpt3, liu2023pre, han2022ptr}. It addresses the domain shift problem, offering a way to adapt source domain input data to the target domain downstream task without parameter tuning of the source model. We use a pixel-level perturbation added around the edges of images as a learnable prompt as in \cite{firstvp} to facilitate task performance.

As the primary role of the learnable prompt is to amend the input image, our prompt enhancer adds a learnable prompt to in-context pairs. Such extended input-output examples will implicitly instruct the frozen model on the desired output and thus narrow the gap between in-context pairs and a query image. Our learnable prompt is agnostic to input, so the same prompt is shared for all in-context pairs of the same task. This means our learnable prompts can be viewed as a task identifier. 

Given the pair $(x, y)$ from the prompt retriever, the prompt enhancer adds to them a learnable prompt $t_{\phi}$ parameterized by $\phi$ to generate $(x', y')$ as
\begin{equation}
  x' = x + \delta t_\phi , \;\; y' = y + \delta t_\phi,
  \label{eq:vp_on_sp}
\end{equation}
where $\delta$ specifies the magnitude of the perturbation. The prompt $t_\phi$ is in the image space. $\phi$ denote the set of pixel around the edges that are learnable via backpropagation, and the other pixels are all zero. 

\subsection{Prediction}

Following \cite{maevqgan}, we adopt the MAE-VQGAN model, in which pre-trained MAE \cite{mae} $E$ generates visual tokens $\hat{z}$ from $(x', y')$ and $x_\text{q}$. The VQGAN \cite{vqgan} decoder $D$, again pre-trained, generates resulting image $\hat{y}_\text{q}$ from $\hat{z}$. 
 
After compiling the in-context pair and the query into a canvas $\hat{c} = (x', y', x_\text{q}, \varnothing)$, $E$ predicts latent visual tokens $\hat{z} = (\hat{z}_1,\dots,\hat{z}_K)$, specifically,
\begin{align}
    \hat{z}_k = \argmax_w E_{kw}(\hat{c}), \label{eq:mae_encoding}
\end{align}
where $\hat{z}_k \in \hat{z}$ is a visual token in the vocabulary $\mathcal{V}$ at spatial position $k$, and $E_{kw}$ gives the probability of $w \in \mathcal{V}$ for $k$. $D$ then generates a label image by 
\begin{align}
    \hat{y}_\text{q} = D(\hat{z}).
\end{align} 
We obtain the prediction for the query $x_\text{q}$ as $\hat{y}_\text{q}$. 

\subsection{Training}

The only learnable parameters in InMeMo are the prompt $t_\phi$. We train it for a specific task on $\mathcal{S}$. The loss is the same as \cite{maevqgan}, while all parameters except for $t_\phi$ are frozen. 

We first randomly choose a pair $(x_\text{q}, y_\text{q})$ as query from $\mathcal{S}$. The InMeMo prediction process from the prompt retriever is then applied to $x_\text{q}$ to compute $\hat{z}$, but the retriever uses $\mathcal{S}\setminus\{(x_\text{q}, y_\text{q})\}$ instead of $\mathcal{S}$. 

The label image $y_\text{q}$ is used for training. We compile the retrieved in-context pair $(x, y)$ and $(x_\text{q}, y_\text{q})$ into a canvas $c = (x, y, x_\text{q}, y_\text{q})$.
The pre-trained VQGAN encoder $F$ associated $D$ gives the ground-truth visual tokens $z$ that reconstruct $y_\text{q}$ with $D$, \ie,
\begin{align}
    z_k = \argmax_w F_{kw}(c),
\end{align}
where $F_{kw}$ again is the probability of $w \in \mathcal{V}$ for position $k$.
The loss $L$ to train our learnable prompt $t_\phi$ is given by
\begin{align}
    L(\phi) = \mathbb{E}[\text{CE}(E_{k}(\hat{c}), z_k)], \label{eq:loss}
\end{align}
where $\text{CE}$ is the cross-entropy loss, $E_{k}(\hat{c}) \in \mathbb{R}^{|\mathcal{V}|}$ is the probabilities of respective tokens in $\mathcal{V}$, and the expectation is computed over all $(x_\text{q}, y_\text{q}) \in \mathcal{S}$ as well as all visual tokens $z_k$ corresponding to $y_\text{q}$ (\ie over the latent visual tokens of $\varnothing$, represented as masked index).

\subsection{Interpretation}

Adding $t_\phi$ to images in a visual prompt as in Eq.~(\ref{eq:vp_on_sp}) translates the distribution of the prompt in a certain direction. Determining $t_\phi$ by Eq.~(\ref{eq:loss}) will encode some ideas about the task described by $\mathcal{S}$ in $\phi$, supplying complementary information that is not fully conveyed by the in-context pair $(x, y)$. We consider that our training roughly aligns the distributions of image patches $\hat{c}$ and $c$ in the latent space before visual token classification with smaller degrees of freedom in $t_\phi$. This can be particularly effective as these distributions are inherently different due to the lack of the ground-truth label image $y_\text{q}$ in the canvas. Therefore, our best expectation is that $\phi$ captures the distribution of $y_\text{q}$ collectively to bring the distribution of prompts closer to the ground-truth prompts (containing ground-truth label $y_\text{q}$). With this, the encoder $E$ will have better access to more plausible visual tokens that decode a label image closer to the ground-truth label.

\section{Experiments}

\begin{table*}[t]
    \centering
    \caption{Performance of the foreground segmentation and single object detection downstream tasks. The best scores in \textit{in-context learning} are highlighted in \textbf{bold}. The baseline scores are based on our reproduction. \textbf{Seg.}~and \textbf{Det.}~stand for the segmentation and single object detection tasks, respectively.}
    \footnotesize
    \begin{tabularx}{\textwidth}{ll>{\centering\arraybackslash}X>{\centering\arraybackslash}X>{\centering\arraybackslash}X>{\centering\arraybackslash}X>{\centering\arraybackslash}X>{\centering\arraybackslash}X}
        \toprule
        & ~ & \multicolumn{5}{c}{\textbf{Seg.} (mIoU $\uparrow$)} & \textbf{Det.} \\
        \cmidrule(lr){3-7} 
        & ~ & Fold-0 & Fold-1 & Fold-2 & Fold-3 & Mean & (mIoU $\uparrow$)  \\ 
        \midrule
        \textit{Meta-learning} 
        &\hspace{5mm}OSLSM \cite{pascal} & 33.60 & 55.30 & 40.90 & 33.50 & 40.80 & -  \\ 
        &\hspace{5mm}co-FCN \cite{cofcn} & 36.70 & 50.60 & 44.90 & 32.40 & 41.10 & -  \\ 
        \midrule
        \textit{In-context learning}
        &\hspace{5mm}\cellcolor{gray!20}Baseline & \cellcolor{gray!20}35.69 & \cellcolor{gray!20}38.25 & \cellcolor{gray!20}35.86 & \cellcolor{gray!20}33.37 & \cellcolor{gray!20}35.79 & \cellcolor{gray!20}28.08  \\ 
        &\hspace{5mm}Random \cite{maevqgan} & 28.66 & 30.21 & 27.81 & 23.55 & 27.56 & 25.45  \\ 
        &\hspace{5mm}UnsupPR \cite{supicl} & 34.75 & 35.92 & 32.41 & 31.16 & 33.56 & 26.84  \\ 
        &\hspace{5mm}SupPR \cite{supicl} & 37.08 & 38.43 & 34.40 & 32.32 & 35.56 & 28.22  \\
        &\hspace{5mm}prompt-SelF \cite{promptself} & \textbf{42.48} & 43.34 & 39.76 & 38.50 & 41.02 & 29.83  \\
        \midrule
        &\hspace{5mm}InMeMo (Ours) & 41.65 & \textbf{47.68} & \textbf{42.43} & \textbf{40.80} & \textbf{43.14} & \textbf{43.21}  \\
        \bottomrule
    \end{tabularx}
    \label{tab:main results}
\end{table*}

\subsection{Experimental Setup}

\paragraph{Datasets and Downstream Tasks.}
We follow the experimental settings of \cite{maevqgan} to evaluate InMeMo. As downstream tasks, we perform foreground segmentation and single object detection. (1) \textbf{Foreground segmentation} aims to extract apparent objects from the query image with the in-context pair. We use the Pascal-5$^i$ dataset \cite{pascal}, which is split into four-fold subsets, each containing five classes. (2) \textbf{Single object detection} evaluates whether a model can capture fine-grained features specified by a coarse-grained bounding box in the in-context pair \cite{promptself}. We conduct experiments on images and bounding boxes from the PASCAL VOC 2012 \cite{everingham2010pascal}. To align with \cite{maevqgan}, we use a subset of the dataset whose samples only contain a single object as our dataset $\mathcal{S}$, ensuring the annotation mask occupies less than 50\% of the entire image for the training set, and 20\% for the test set.

\paragraph{Methods for comparison.}
All experiments use MAE-VQGAN \cite{maevqgan} as the pre-trained large-scale vision model. InMeMo is compared against the SOTA methods of visual ICL (\ie, Random \cite{maevqgan}, UnsupPR \cite{supicl}, SupPR \cite{supicl}, and prompt-SelF \cite{promptself}) as well as few-shot segmentation derived from meta-learning (\ie, OSLSM \cite{pascal} and co-FCN \cite{cofcn}). Our baseline is pixel-level retrieval \cite{promptself} but without the learnable prompt.


\paragraph{Implementation details.}
For foreground segmentation, we train InMeMo for each fold of the training set separately, meaning each fold is viewed as a task, and a learnable prompt is obtained for each fold. For single object detection, we train InMeMo on the whole training set by retrieving in-context pairs from the training set. For testing, each image in the test set will be considered as a query image to retrieve an in-context pair from the training set.

We resized the image size to 224 $\times$ 224 for the prompt enhancer. A learnable prompt occupies 30 pixels from each edge; therefore, $\phi$ contains $(224^2-(224-2\times 30)^2)\times 3$ parameters. The images are then resized to 111 $\times$ 111 to create the canvas. An in-context pair $(x', y')$ with a learnable prompt, a query image $x_\text{q}$, and an empty image $\varnothing$ are arranged at top-left, top-right, bottom-left, and bottom-right, respectively, following the default arrangement of \cite{maevqgan}. We set $\delta$ to 1. InMeMo is implemented using PyTorch and trained for 100 epochs with Adam \cite{adam}. We initiate training with a learning rate of 40, which decays based on the cosine annealing warm restarts scheduler. A notable advantage of InMeMo is its efficiency\textemdash this training operates on a single NVIDIA GeForce RTX 4090 with a batch size of 32.

\subsection{Comparison with State-of-the-Art}

We compared InMeMo with prior visual ICL methods and meta-learning-based few-shot learning methods in \cref{tab:main results}. Our analysis reveals that InMeMo achieved the SOTA, surpassing the previous SOTA in both downstream tasks, particularly on single object detection. Apparently, it significantly outperformed the baseline. Our method also outperformed the meta-learning-based methods on some folds and on average. This highlights the efficacy of integrating the learnable prompt into visual ICL.

More specifically, for the foreground segmentation task, we observe that while InMeMo does not achieve the best score on Fold-0, it nonetheless considerably exceeds the baseline. Prompt-SelF's performance could be affected by the bagging effect. That is, prompt-SelF is applied to eight different arrangements of images in the canvas and fuses the results, thereby harnessing the latent expertise of the large-scale vision model. In contrast, InMeMo runs inference for a single query only once. Bagging can be an interesting tweak to improve the performance without extensive efforts, but still, we emphasize the significant gain of InMeMo by itself.  Notably, InMeMo showcases outstanding proficiency in the single object detection task, surpassing the prevailing SOTA by a margin of 13.38 points. This performance gain, demonstrates the exceptional ability of InMeMo to capture fine-grained features in detecting small objects within images.

\begin{table*}[t]
    \centering
    \caption{Domain shift analysis on InMeMo. \textit{Pascal $\rightarrow$ Pascal} means in-context pairs and query images both source from PASCAL (as with \cref{tab:main results}). \textit{COCO $\rightarrow$ Pascal} indicates that in-context pairs are from COCO and query images are from PASCAL. The baseline scores are our reproduction.}
    \footnotesize
    \begin{tabularx}{\textwidth}{>{\centering\arraybackslash}m{2cm}>{\centering\arraybackslash}m{4cm}>{\centering\arraybackslash}X>{\centering\arraybackslash}X>{\centering\arraybackslash}X>{\centering\arraybackslash}X>{\centering\arraybackslash}X}
        \toprule
        & Method & Fold-0 & Fold-1 & Fold-2 & Fold-3 & Means  \\
        \midrule
        \textit{Pascal $\rightarrow$ Pascal} & \cellcolor{gray!20}Baseline & \cellcolor{gray!20}35.69 & \cellcolor{gray!20}38.25 & \cellcolor{gray!20}35.86 & \cellcolor{gray!20}33.37 & \cellcolor{gray!20}35.79 \\
        & InMeMo & 41.65 & 47.68 & 42.43 & 40.80 & 43.14 \\
        \midrule
        \textit{COCO $\rightarrow$ Pascal} & \cellcolor{gray!20}Baseline & \cellcolor{gray!20}33.83 & \cellcolor{gray!20}36.11 & \cellcolor{gray!20}32.89 & \cellcolor{gray!20}30.64 & \cellcolor{gray!20}33.37 \\
        & InMeMo & 38.74 & 43.82 & 40.45 & 37.12 & 40.03  \\
        \bottomrule
    \end{tabularx}
    \label{tab:ds}
\end{table*}

These results shed light on our direct and efficient approach. By amending in-context pairs, we can effectively harness the learnable prompt to improve performance in visual ICL. Moreover, InMeMo stands out with its lightweight nature, using only 69,840 additional parameters and demanding minimal training resources. The shared pixel-level learnable prompts of InMeMo hold the potential to pave the way toward even more efficient and effective visual ICL.

\subsection{Domain Shift Analysis}

Real-world applications often exhibit domain shifts from dataset $\mathcal{S}$, leading to discrepancies in model performance in comparison with in-domain evaluation due to differing distributions. Such domain shifts can be observed across datasets, and the resulting performance disparities among datasets can serve as a benchmark for evaluating model robustness \cite{domaing}. To assess InMeMo's sensitivity to domain shift, we employ the COCO dataset \cite{coco} for inference, following the same setting as in \cite{supicl}. The COCO dataset is divided into four subsets, each of which mirrors the categories of Pascal-5$^i$, denoted as COCO-5$^i$ \cite{supicl}. We source the in-context pair from COCO-5$^i$ and obtain the query image from the validation set of Pascal-5$^i$, consistent with \cite{promptself}. This specific configuration is termed as \textit{COCO $\rightarrow$ Pascal}.

\Cref{tab:ds} summarizes our domain shift evaluation results. For the \textit{COCO $\rightarrow$ Pascal} configuration, the baseline marks a drop in mIoU score by 2.42. In contrast, InMeMo hits 40.03\%, reflecting a drop of 3.11. The gap between them is 0.69, indicating that InMeMo is robust against domain shift. Consequently, visual ICL with the learnable prompt has the potential to be transferable, making InMeMo reliable for various real-world applications.

\begin{table}[t]
    \centering
    \caption{Segmentation performance for some combinations of images in a canvas to which the learnable prompt are added. \texttt{I}, \texttt{L}, and \texttt{Q} means in-context image, in-context label image, and query image, respectively.}
    \resizebox{\columnwidth}{!}{
    \begin{tabular}{lccccc}
        \toprule
        ~ & Fold-0 & Fold-1 & Fold-2 & Fold-3 & Mean  \\ 
        \midrule
        \rowcolor{gray!20} Baseline & 35.69 & 38.25 & 35.86 & 33.37 & 35.79  \\
        \midrule
        prompt-SelF \cite{promptself} & 42.48 & 43.34 & 39.76 & 38.50 & 41.02  \\
        \midrule
        \multicolumn{6}{l}{Combination (InMeMo variant)} \\
        \hspace{5mm}\texttt{I} & \textbf{42.57} & 47.08 & 41.60 & 39.44 & 42.67  \\ 
        \hspace{5mm}\texttt{Q} & 39.56 & 44.57 & 41.40 & 38.06 & 40.90  \\ 
        \hspace{5mm}\texttt{I} \& \texttt{Q} & 38.31 & 44.37 & 39.98 & 37.80 & 40.12  \\
        \hspace{5mm}\texttt{I} \& \texttt{L} (InMeMo) & 41.65 & \textbf{47.68} & \textbf{42.43} & \textbf{40.80} & \textbf{43.14}  \\        \hspace{5mm}\texttt{I}, \texttt{L} \& \texttt{Q} & 39.84 & 43.49 & 35.58 & 27.39 & 36.58  \\

        \bottomrule
    \end{tabular}
    }
    \label{tab:location vp}
\end{table}

\subsection{More Analysis on InMeMo}

This section further investigates the capabilities of InMeMo through a series of experiments, primarily focusing on the foreground segmentation task. 

\begin{figure*}[t]
  \centering
   \includegraphics[width=1\linewidth]{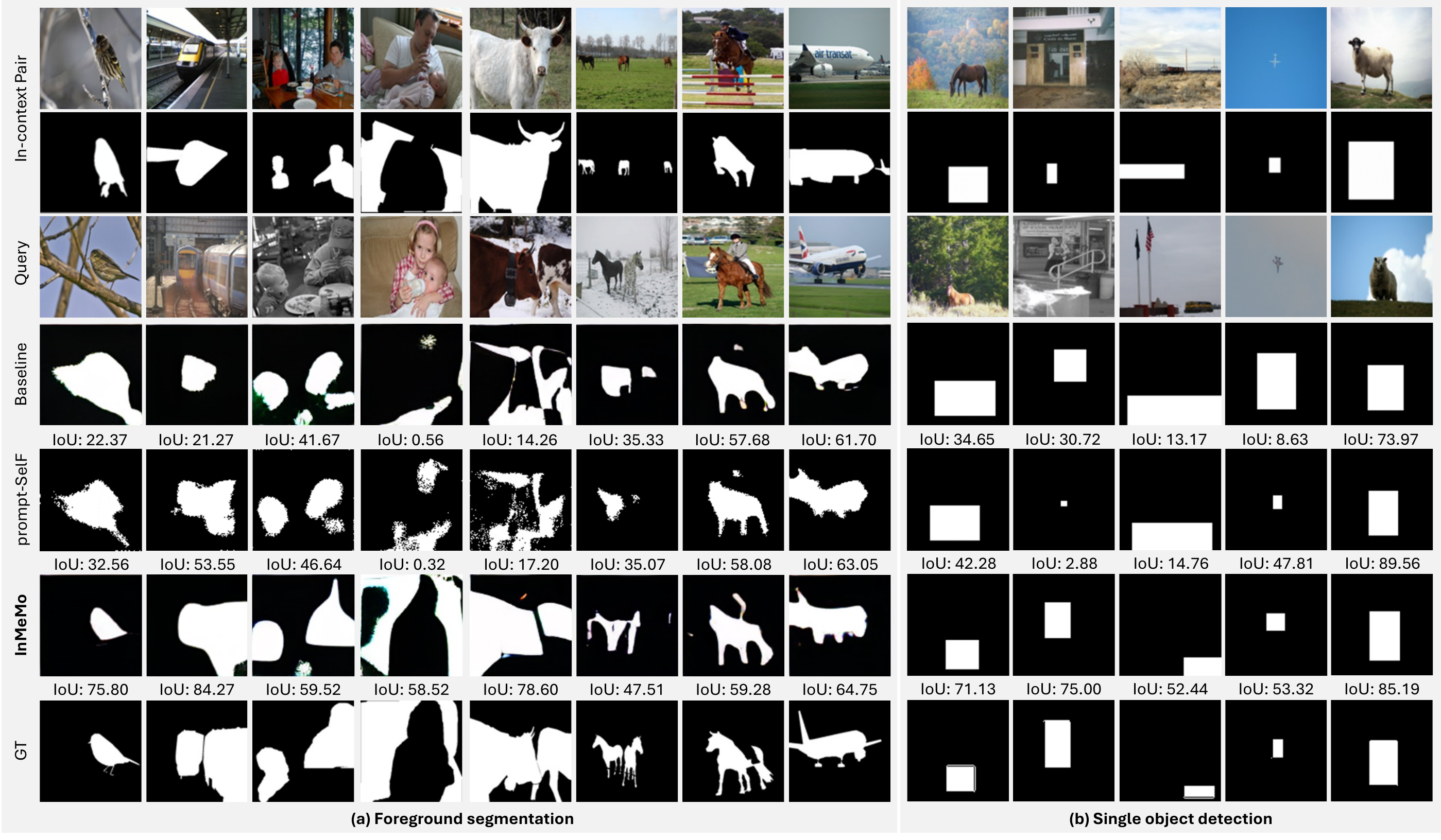}
   \caption{Some examples of baseline, prompt-SelF, and our InMeMo over the two downstream tasks: \textbf{(a)} Foreground segmentation and \textbf{(b)} Single-object detection. In each task, the upper two rows are in-context pairs, and the third row is the query image. We arrange rows from top to bottom with the order of the baseline, prompt-SelF, \textbf{InMeMo}, and the ground-truth label (GT). InMeMo can lead the visual ICL to capture detailed features and overcome inconsistency between in-context and query images. Moreover, InMeMo behaves like it neglects poor-quality in-context pairs, which is another strong advantage when the prompt retriever cannot find a similar in-context image. More examples can be found in our supplementary material.}
   \label{fig:visual_examples}
\end{figure*}

\paragraph{Qualitative comparison.} We qualitatively compare InMeMo with the baseline, prompt-SelF\footnote{We reproduced the prompt-SelF to generate visual examples.} \cite{promptself}, and the ground-truth (GT) label using examples for both foreground segmentation and single object detection tasks, which are shown in \cref{fig:visual_examples}. For the foreground segmentation task (\cref{fig:visual_examples}(a)), InMeMo produces details faithful to the ground-truth label images. Interestingly, InMeMo remains robust against variations, including when provided with an achromatic image or when a significant color disparity exists between the in-context and the query images. Moreover, the InMeMo appears resistant to variations in foreground size (the fourth column from the right), and it distinguishes the background in the query image. However, when the in-context and query images closely align in terms of their features (\eg, semantics, viewpoints, sizes, poses \cite{supicl}), InMeMo's performance matches that of the baseline and prompt-SelF.

In the single object detection task (\cref{fig:visual_examples}(b)), InMeMo consistently displays its detail-oriented nature and is unfazed by color variations or object size differences in the in-context pair. Particularly notable is its competency in scenarios where the presence of the foreground in the in-context pair is minimal. Nevertheless, akin to the segmentation task, when the in-context and query images bear a strong resemblance, InMeMo mirrors the performance of the prompt-SelF.

\paragraph{Which images should the learnable prompt be added?} Our recommendation leans towards introducing the learnable prompt only to in-context pairs, which seems pivotal in enhancing visual ICL's efficacy across tasks, given the guiding nature of in-context pairs. To discern the potential impacts of different combinations of images to which the learnable prompt is added, we assessed five InMeMo variants: only \textit{in-context image} (\texttt{I}), only \textit{query image} (\texttt{Q}), both \textit{in-context image} and \textit{query image} (\texttt{I} \& \texttt{Q}), \textit{in-context image} and \textit{in-context label image} (\texttt{I} \& \texttt{L}, identical to InMeMo), and \textit{in-context image}, \textit{in-context label image}, and \textit{query image} (\texttt{I}, \texttt{L}, \& \texttt{Q}).

The scores of these combinations are summarized in \cref{tab:location vp}. We found that the learnable prompt, irrespective of location, improves the visual ICL performance. Surprisingly, adding the prompt to in-context images produced suboptimal performance among the InMeMo variants but outperformed prompt-SelF. This indicates that the learnable prompt effectively improves the quality of the in-context pair and can enhance the visual ICL performance.

We also found that the performance is less effective when we add the learnable prompt to in-context and query images (\texttt{I} \& \texttt{Q}) than when we add it to only one image (\texttt{I}, \texttt{Q}). This can be attributed to the agnostic learnable prompt. When the identical prompt is added to both in-context and query images, the model struggles to narrow the gap between them effectively. Similarly, this underlying factor leads to compromised performance in the \texttt{I}, \texttt{L}, \& \texttt{Q} configuration, with a particularly notable performance reduction in the more challenging Fold-3.

\begin{figure}[t]
  \centering
   \includegraphics[width=1\linewidth]{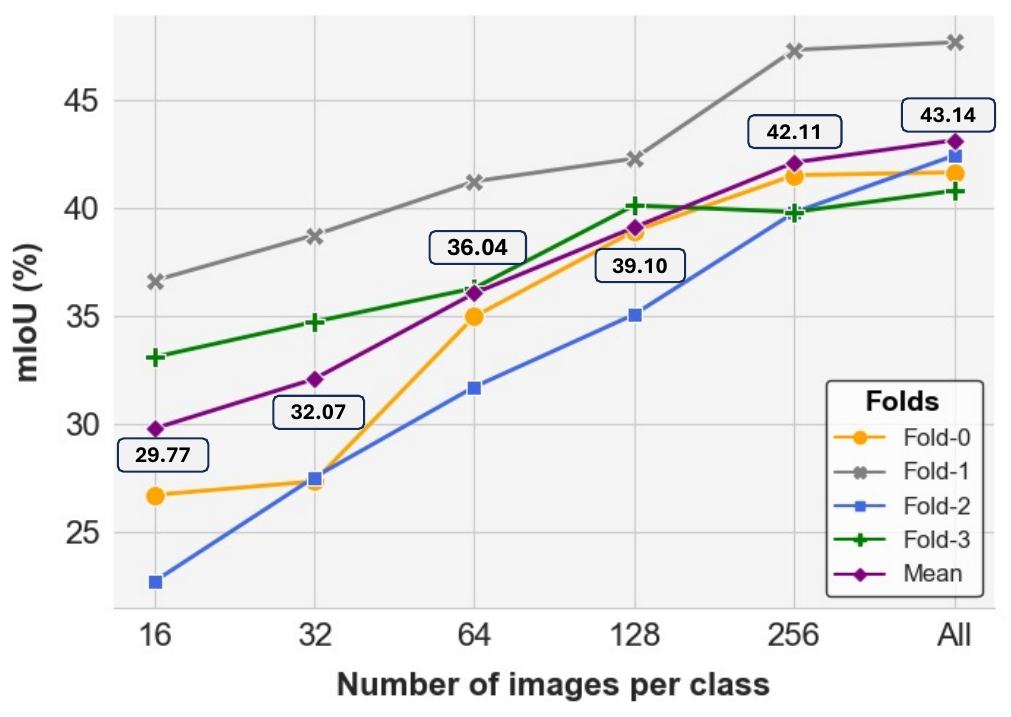}
   \caption{The performance of InMeMo in mIoU of each fold for the number of images per class in $\mathcal{S}$. All means to use all images in the training set. We annotate the scores of \textcolor{Plum}{Mean} in the figure.}
   \label{fig:fsl_result}
\end{figure}

\paragraph{Is InMeMo performance sensitive to the dataset size?} Given the efficiency and simplicity of the learnable prompt, we sought to elucidate the relationship between the volume of the dataset $\mathcal{S}$ and the performance of InMeMo. We conducted experiments for each fold, randomly picking 16, 32, 64, 128, and 256 images from each class to compose $\mathcal{S}$. \Cref{fig:fsl_result} depicts the relationship.

Our empirical results suggest that the overall performance (represented as \textcolor{Plum}{Mean}) surpasses the baseline score (35.79\%) when using at least 64 images per class (36.04\%). The performance tends to improve as the number of images increases. Specifically, for Fold-1, which is comparatively easy, InMeMo achieves the mIoU accuracy of 36.63\% with only 16 images and consistently outperforms the other folds. Fold-0 substantially increases accuracy, starting from 32 images, saturated at 256. Fold-2 consistently shows a significant improvement as the number of images increases. In Fold-3, there is a considerable increase from 64 to 128 images, after which the score becomes saturated and only gradually increases when all images are used. In general, for easier folds, InMeMo requires fewer images; however, when dealing with intricate scenes, increasing the dataset size can enhance the performance of InMeMo.

\begin{figure}[t]
  \centering
   \includegraphics[width=1\linewidth]{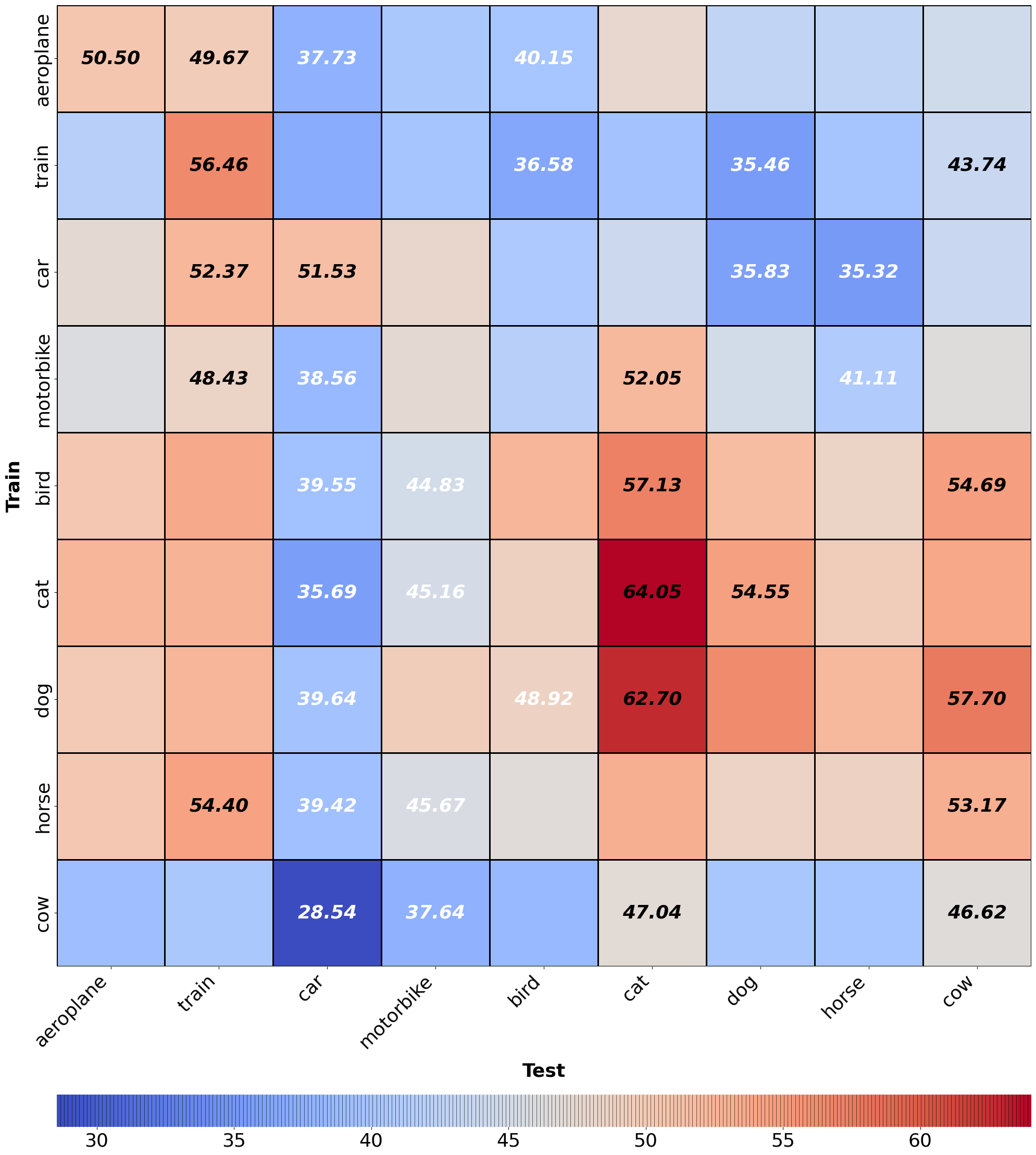}
   \caption{Inter- and intra-class generalization performance in mIoU. The horizontal and vertical axes are the classes used for prediction and training, respectively. The diagonal elements show intra-class performance. Each row shows the two largest and smallest scores in black and white. The nine classes are arranged to form \textit{\small{}transportation} and \textit{\small{}animal} super-classes. The supplementary material shows the scores for all possible pairs.}
   \label{fig:cls_base}
\end{figure}

\paragraph{Inter-class generalizability of InMeMo.} We have demonstrated that InMeMo works well on poor datasets and are curious about its generalizability to unseen classes not included in the dataset $\mathcal{S}$. For this, we train a learnable prompt for each of the 20 classes. Specifically, let $\mathcal{S}_\omega$ denote the subset of images and label images in Pascal-5$^i$ for class $\omega$. InMeMo training uses an image in $\mathcal{S}_\omega$ as a query. It also uses a pair of an image and a label image in $\mathcal{S}_\omega$ as an in-context pair. We then run predictions using images in $\mathcal{S}_{\omega'}$ as queries and in-context pairs, where $\omega' \neq \omega$, for measuring the inter-class generalizability. As different classes have varying levels of difficulties, we only show the classes whose \textit{intra-class} performance is higher than the mean mIoU score (43.14\%) in \cref{tab:main results}. Our supplementary material shows full results. We discovered that the \texttt{\small{}bus} and \texttt{\small{}sheep} are the most \textit{general} classes, meaning that prompts trained on different classes yield a high accuracy (mIoU above 50\%) on these two classes. In contrast, \texttt{\small{}person} is the least generalizable class, performing poorly on all other classes. We excluded these three classes as well, ending up with nine classes. The inter-class (as well as intra-class) scores are shown in \cref{fig:cls_base}.

The figure indicates that intra-class scores are not always the best among all other classes. The \texttt{\small{}transportation} super-class, we can see strong generalizability between the classes in it (\texttt{\small{}aeroplane}-\texttt{\small{}train}, \texttt{\small{}car}-\texttt{\small{}train}, and \texttt{\small{}motorbike}-\texttt{\small{}train}), whereas the \texttt{\small{}transportation} classes typically have lower scores with the \texttt{\small{}animals} classes like \texttt{\small{}dog} and \texttt{\small{}horse}. Within \texttt{\small{}animals}, the classes usually show strong generalizability except for \texttt{\small{}cow}, but they do not generalize to the \texttt{\small{}transportation} classes. We can also identify some exceptions between different classes, such as \texttt{\small{}aeroplane} having a weak generalizability with \texttt{\small{}car} and \texttt{\small{}horse} having a strong generalizability with \texttt{\small{}train}. We think this is due to the similarity of their label images (\eg, \texttt{\small{}train} and \texttt{\small{}cow} often occupy a larger region of the image) and the class-specific difficulty (\eg, the learnable prompt trained for \texttt{\small{}cow} does not generalize in most cases).

The mean mIoU score over all pairs of 20 classes in the supplementary material is 34.32\%. This score is comparable to most methods in \cref{tab:main results}, but suffers from a significant drop from InMeMo's mean score over all folds. This implies the importance of tuning the learnable prompt for target tasks.

\section{Conclusion}

InMeMo shows SOTA performance on the two downstream tasks by incorporating a learnable prompt to in-context pairs, a lightweight tool to facilitate visual ICL. The learnable prompt enables the visual ICL to reconstruct more fine-grained details in predictions and overcome the interference caused by low-quality in-context pairs that are not sufficiently similar to query images. We also showed that InMeMo is robust against domain shift (\eg, from the Pascal dataset to the COCO dataset). \textbf{Limitations.} InMeMo requires a minimum of 64 images per class to achieve competitive performance compared with our baseline. Also, a learnable prompt for a certain class does not generalize to other classes. Therefore, the learnable prompt dedicated to the target task is the key to better performance.

\vspace{-1mm}
\paragraph{Acknowledgements}

This work was partly supported by JSPS KAKENHI Grant No.~JP23H00497, JST CREST Grant No.~JPMJCR20D3, and FOREST Grant No.~JPMJFR216O.



{\small
\bibliographystyle{ieee_fullname}
\bibliography{egbib}
}

\newpage

\appendix
\renewcommand\thesection{\Alph{section}}
\section{Complete Table of Inter- and Intra-class Generalizability}

\begin{figure*}[t]
  \centering
   \includegraphics[width=1\linewidth]{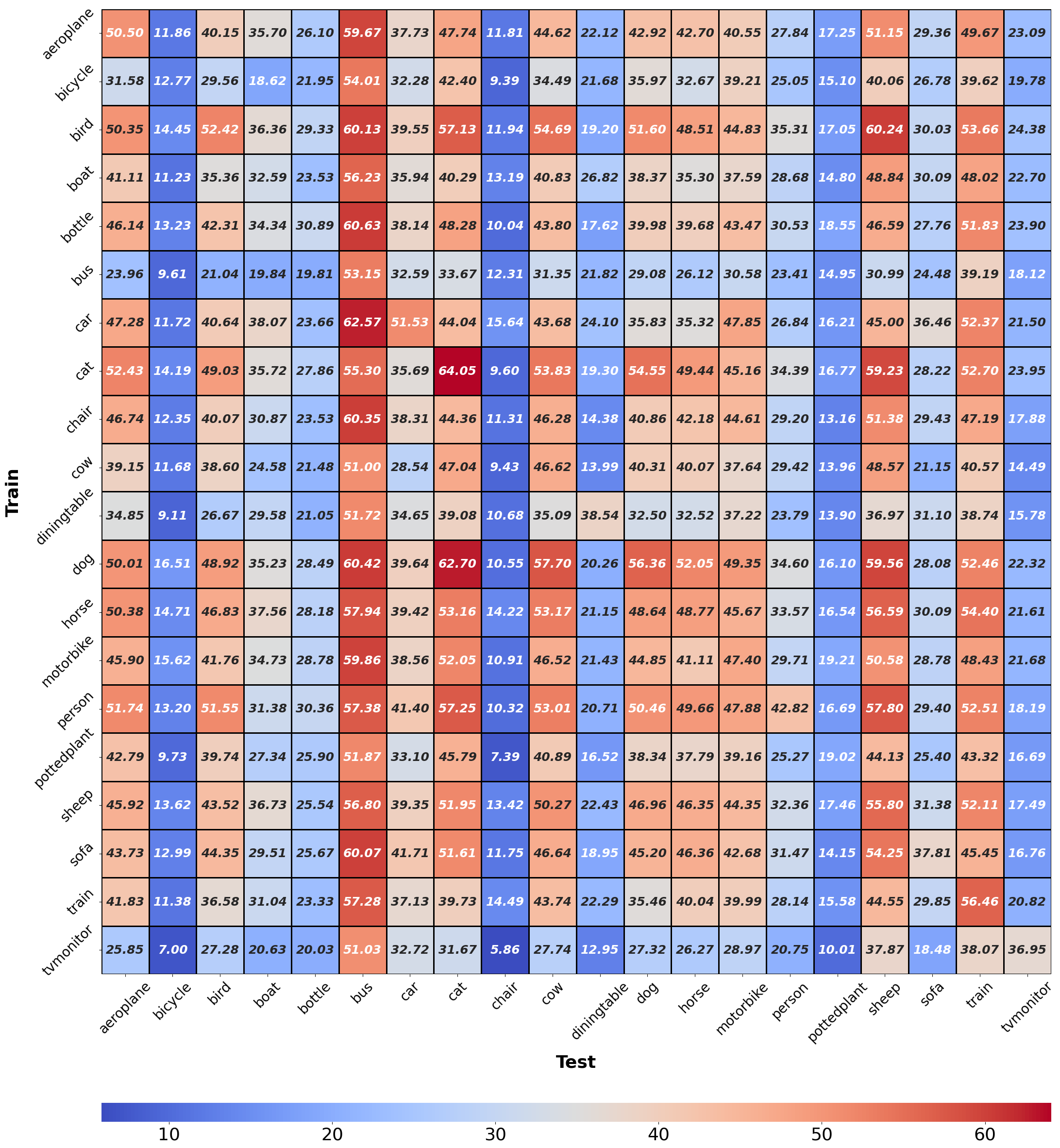}
   \caption{The complete Inter- and intra-class generalization performance in mIoU. The horizontal and vertical axes are the classes used for prediction and training, respectively. The diagonal elements show intra-class performance.}
   \label{fig:full_cls_base}
\end{figure*}

Due to the differences in dataset difficulty, we established criteria and filtered categories for presentation in the main manuscript. We display the complete result in \cref{fig:full_cls_base} to show the inter- and intra-class generalizability of InMeMo's performance in mIoU on Psacal-5$^i$ dataset \cite{pascal}.

According to the column (class-level generalizability of test set in $\mathcal{S}_{\omega'}$), we found \texttt{\small{}bus} and \texttt{\small{}sheep} are \textit{general} classes. The learnable prompt, trained on any classes, performed well on the test classes of \texttt{\small{}bus} (56.87 $\pm$ 3.57) and \texttt{\small{}sheep} (49.01 $\pm$ 8.07) in visual ICL. Regarding other classes, we found that some classes are difficult tasks for visual ICL, leading to poor performance, such as \texttt{\small{}bicycle} (12.35 $\pm$ 2.26), \texttt{\small{}bottle} (25.27 $\pm$ 3.36), \texttt{\small{}chair} (11.21 $\pm$ 2.30), \texttt{\small{}diningtable} (20.81 $\pm$ 5.28), \texttt{\small{}pottedplant} (15.82 $\pm$ 2.13), \texttt{\small{}sofa} (28.71 $\pm$ 4.24), and \texttt{\small{}tvmonitor} (20.90 $\pm$ 4.68). The \texttt{\small{}person} is the least generalizable class, where the learnable prompt trained on \texttt{\small{}person} is most effective on the test set of \texttt{\small{}person}, while that trained on other classes perform poorly.

The mean mIoU score for all 20 class pairs is 34.32\%, which suffers from a significant drop from InMeMo's mean score over all folds. This indicates the need to adjust the learnable prompt for target tasks.

\section{More Results of Domain Shift Analysis}

To assess the robustness of incorporating the learnable prompt at various variants regarding the domain shift issues, we present a comprehensive comparison in \cref{tab:main ds results}. This analysis was carried out on the COCO-5$^i$ dataset \cite{supicl}, with identical settings mentioned in the main manuscript.

Overall, the addition of the learnable prompt at each variant is robust. Specifically, the results for the drop score of \textit{Means} on different variants are as follows: baseline (2.42), \texttt{I} (2.74), \texttt{Q} (2.51), \texttt{I} \& \texttt{Q} (2.54), \texttt{I} \& \texttt{L} (\ie InMeMo, 3.11), \texttt{I}, \texttt{L} \& \texttt{Q} (1.5).

We observed that the rankings for performance on \textit{Means} did not change. On the relatively \textit{easy} splits (based on baseline) on Fold-0 and Fold-1, \texttt{I} achieved the best performance of all variants (40.55\% \& 44.53\%). On the more \textit{hard} splits Fold-2 and Fold-3, InMeMo performed better (40.45\% \& 37.12\%). The \texttt{I} meets a relatively minor drop compared to InMeMo, which means giving the learnable prompt to in-context label images can improve the overall performance but sacrifice the robustness of domain shift issues. In variant \texttt{I}, \texttt{L} \& \texttt{Q}, it exhibits a minimal decrease in domain shift issue, making it the most robust among all variants. Consequently, compared to the baseline, all variants demonstrate robustness to domain shift issues. It would be an intriguing research direction to explore the development of learnable prompts that are more robust for visual ICL.

\begin{table*}[t]
    \centering
    \caption{The results of domain shift analysis on all the variants of InMeMo. \textit{Pascal $\rightarrow$ Pascal} means in-context pairs and query images both source from PASCAL. \textit{COCO $\rightarrow$ Pascal} indicates that in-context pairs are from COCO-5$^i$ and query images are from Pascal-5$^i$. The baseline scores are our reproduction of pixel-level retrieval in \cite{promptself}. The best results in each column in \textit{COCO $\rightarrow$ Pascal} are represented in \textbf{bold}.}
    \footnotesize
    \begin{tabularx}{\textwidth}{>{\centering\arraybackslash}m{2cm}>{\raggedright\arraybackslash}m{4cm}>{\centering\arraybackslash}X>{\centering\arraybackslash}X>{\centering\arraybackslash}X>{\centering\arraybackslash}X>{\centering\arraybackslash}X}
        \toprule
        ~ & \hspace{5mm}Combination & Fold-0 & Fold-1 & Fold-2 & Fold-3 & Means  \\
        \midrule
        \multirow{7}{*}{\textit{Pascal $\rightarrow$ Pascal}} &
        \hspace{5mm} \cellcolor{gray!20}Baseline & \cellcolor{gray!20}35.69 & \cellcolor{gray!20}38.25 & \cellcolor{gray!20}35.86 & \cellcolor{gray!20}33.37 & \cellcolor{gray!20}35.79 \\
        & \hspace{5mm}\texttt{I} & 42.57 & 47.08 & 41.60 & 39.44 & 42.67  \\ 
        & \hspace{5mm}\texttt{Q} & 39.56 & 44.57 & 41.40 & 38.06 & 40.90  \\ 
        & \hspace{5mm}\texttt{I} \& \texttt{Q} & 38.31 & 44.37 & 39.98 & 37.80 & 40.12  \\
        & \hspace{5mm}\texttt{I} \& \texttt{L} (InMeMo) & 41.65 & 47.68 & 42.43 & 40.80 & 43.14 \\
        & \hspace{5mm}\texttt{I}, \texttt{L} \& \texttt{Q} & 39.84 & 43.49 & 35.58 & 27.39 & 36.58  \\
        \midrule
        \multirow{7}{*}{\textit{COCO $\rightarrow$ Pascal}} &
        \hspace{5mm} \cellcolor{gray!20}Baseline & \cellcolor{gray!20}33.83 & \cellcolor{gray!20}36.11 & \cellcolor{gray!20}32.89 & \cellcolor{gray!20}30.64 & \cellcolor{gray!20}33.37 \\
        & \hspace{5mm}\texttt{I} & \textbf{40.55} & \textbf{44.53} & 38.62 & 36.01 & 39.93  \\ 
        & \hspace{5mm}\texttt{Q} & 37.26 & 42.40 & 39.33 & 34.56 & 38.39  \\ 
        & \hspace{5mm}\texttt{I} \& \texttt{Q} & 35.70 & 41.96 & 37.58 & 35.06 & 37.58  \\
        & \hspace{5mm}\texttt{I} \& \texttt{L} (InMeMo) & 38.74 & 43.82 & \textbf{40.45} & \textbf{37.12} & \textbf{40.03}  \\
        & \hspace{5mm}\texttt{I}, \texttt{L} \& \texttt{Q} & 38.67 & 41.86 & 33.33 & 26.44 & 35.08  \\
        \bottomrule
    \end{tabularx}
    \label{tab:main ds results}
\end{table*}

\section{The Padding Size of Prompt Enhancer $t_\phi$}

We configured the padding size of InMeMo to 30 followed \cite{firstvp}, but were curious about the impact of various padding sizes on InMeMo's performance. Consequently, we conducted a thorough analysis to assess the effects of altering the padding size as shown in \cref{tab:padding size}.

We found that our InMeMo performed the best regarding Mean in all padding size settings. Increasing the padding size from 10 to 30 resulted in improved performance for InMeMo. In contrast, when we increase the padding size from 30 to 60, we found a decrease in InMeMo's performance.

When the padding size is set to 20, sub-optimal results are obtained on the Mean, while the best performance is achieved on Fold-1 and Fold-2 (48.11\% and 42.68\%, respectively). The performances of InMeMo at settings 20 and 30 achieved similar results. However, increasing the padding size from 30 to 40 led to a significant decrease in performance (from 43.14\% to 26.90\%). Conversely, there was only a relatively slight decrease from 40 to 60. We claim that setting the padding size to 40 causes the learnable prompt to cover much more original information or distribution of the in-context pair, resulting in a considerable loss of performance. Therefore, a padding size of 30 is optimal for InMeMo. Larger padding sizes with additional parameters would result in inferior performance in our proposed InMeMo model.

\begin{table}[t]
    \centering
    \caption{The mIoU scores for varying padding sizes of prompt enhancer $t_\phi$. The Para. represents the number of tunable parameters. The highest score in each fold is marked in \textbf{bold}.}
    \resizebox{\columnwidth}{!}{
    \begin{tabular}{lcccccc}
        \toprule
        Padding size & Para. & Fold-0 & Fold-1 & Fold-2 & Fold-3 & Mean  \\ 
        \midrule
        \hspace{5mm} 10 & 25,680 & 40.19 & 46.16 & 40.87 & 40.06 &  41.82  \\
        \hspace{5mm} 20 & 48,960 & 40.82 & \textbf{48.11} & \textbf{42.68} & 39.12 &  42.68  \\
        \hspace{6mm}\cellcolor{gray!20}30 (InMeMo) & \cellcolor{gray!20}69,840 & \cellcolor{gray!20}\textbf{41.65} & \cellcolor{gray!20}47.68 & \cellcolor{gray!20}42.43 & \cellcolor{gray!20}\textbf{40.80} & \cellcolor{gray!20}\textbf{43.14}  \\ 
        \hspace{5mm} 40 & 88,320 & 24.12 & 29.77 & 27.60 & 26.09 & 26.90  \\ 
        \hspace{5mm} 50 & 104,400 & 20.55 & 28.83 & 24.04 & 27.34 & 25.19  \\
        \hspace{5mm} 60 & 118,080 & 18.73 & 28.50 & 24.78 & 27.51 & 24.88  \\
        \bottomrule
    \end{tabular}
    }
    \label{tab:padding size}
\end{table}

\section{More Visual Examples}

In this section, we illustrate more visual examples from different folds. The main paper only presents a few specific examples, so we want to provide more accessible and intuitive examples of our InMeMo.

\subsection{Foreground segmentation}

\begin{figure*}[t]
  \centering
   \includegraphics[width=0.99\linewidth]{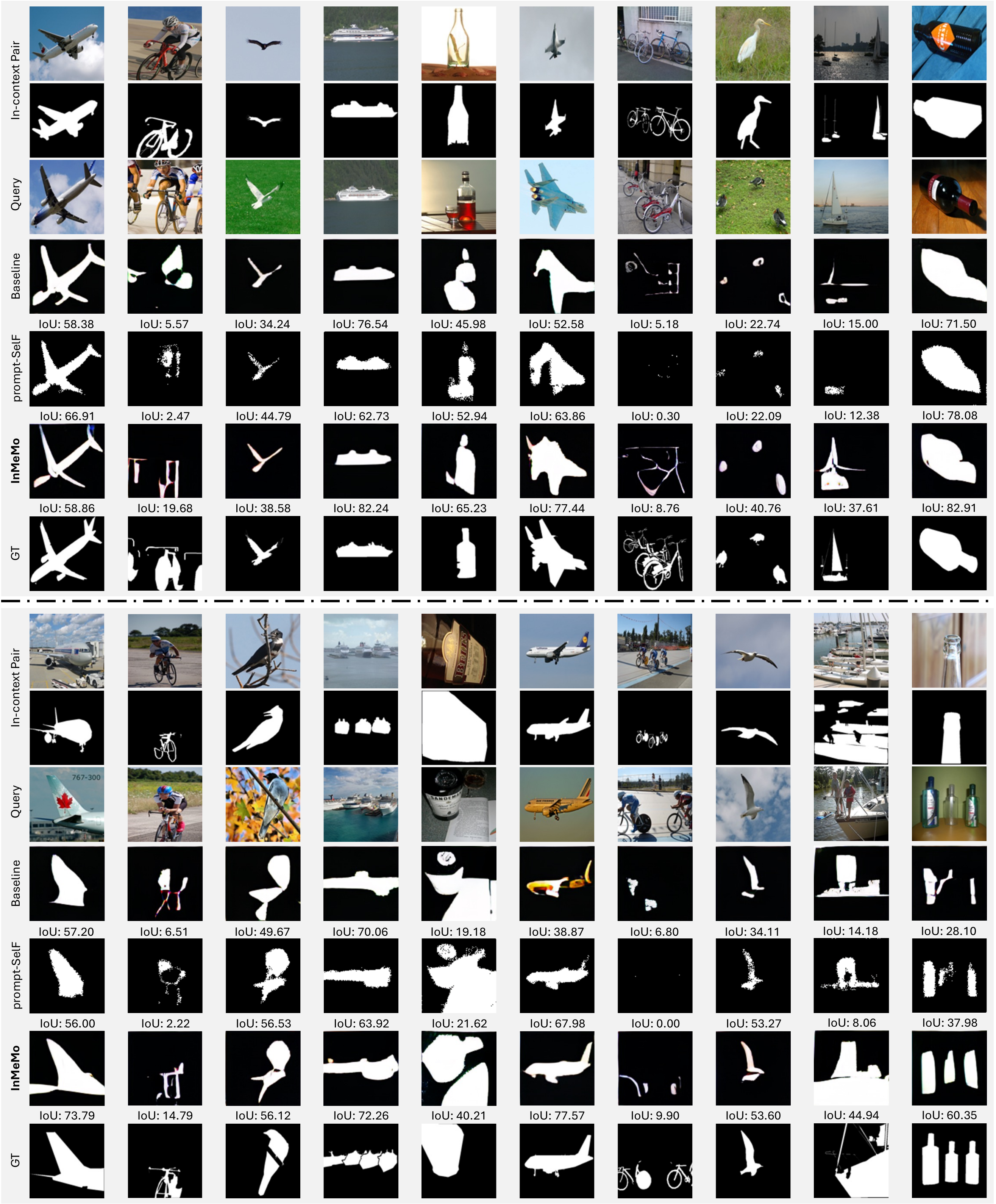}
   \caption{The visual examples in Fold-0. Each column represents a different class. We arrange them in the order of in-context pair, query image, baseline, prompt-SelF, \textbf{InMeMo}, and the ground-truth label (GT). We show all five classes in this fold: \textit{aeroplane}, \textit{bicycle}, \textit{bird}, \textit{boat}, \textit{bottle}.}
   \label{fig:fold0}
\end{figure*}

\begin{figure*}[t]
  \centering
   \includegraphics[width=1\linewidth]{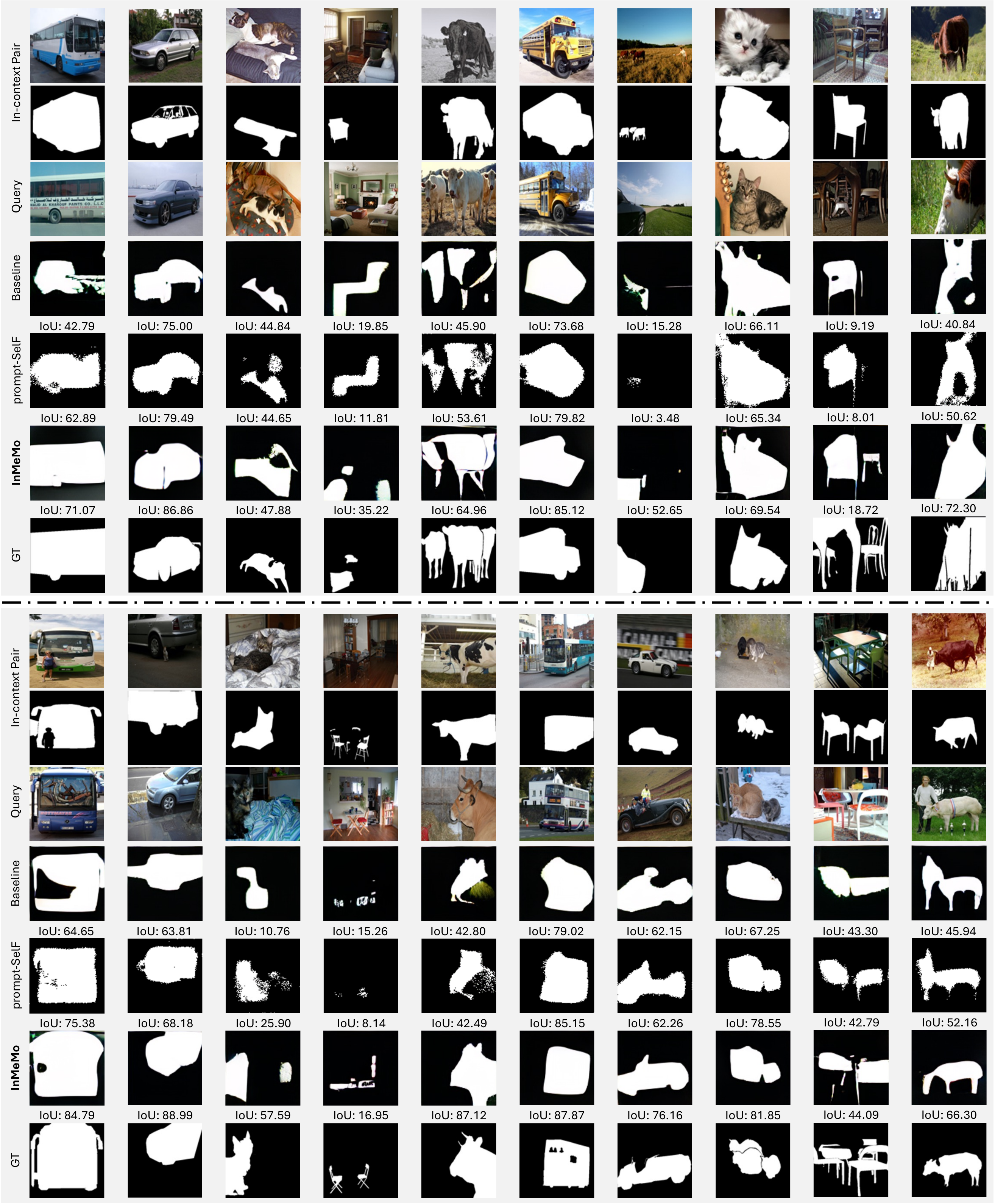}
   \caption{The visual examples in Fold-1: \textit{bus}, \textit{car}, \textit{cat}, \textit{chair}, \textit{cow}.}
   \label{fig:fold1}
\end{figure*}

\begin{figure*}[t]
  \centering
   \includegraphics[width=1\linewidth]{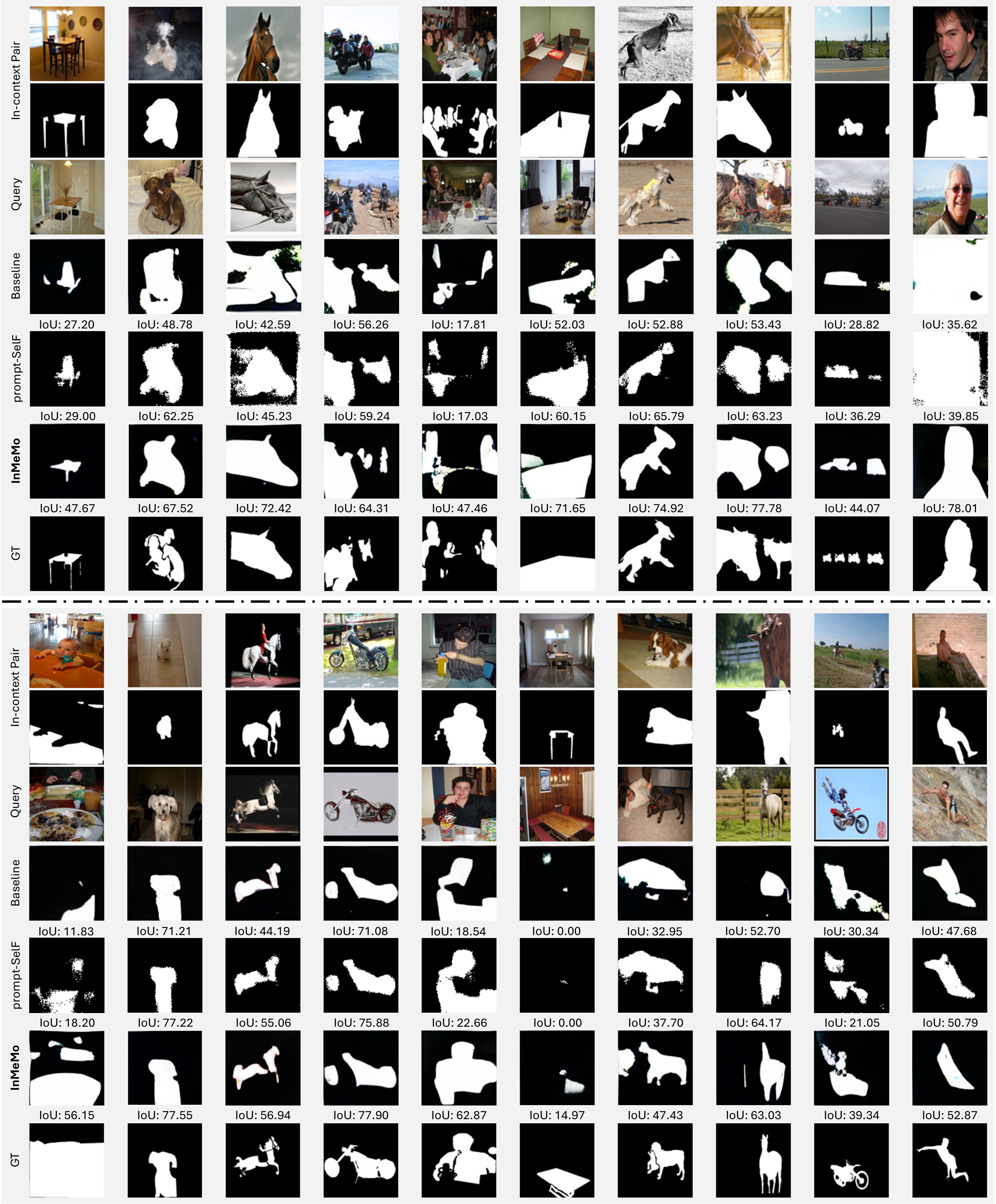}
   \caption{The visual examples in Fold-2: \textit{dinningtable}, \textit{dog}, \textit{horse}, \textit{motorbike}, \textit{person}.}
   \label{fig:fold2}
\end{figure*}

\begin{figure*}[t]
  \centering
   \includegraphics[width=1\linewidth]{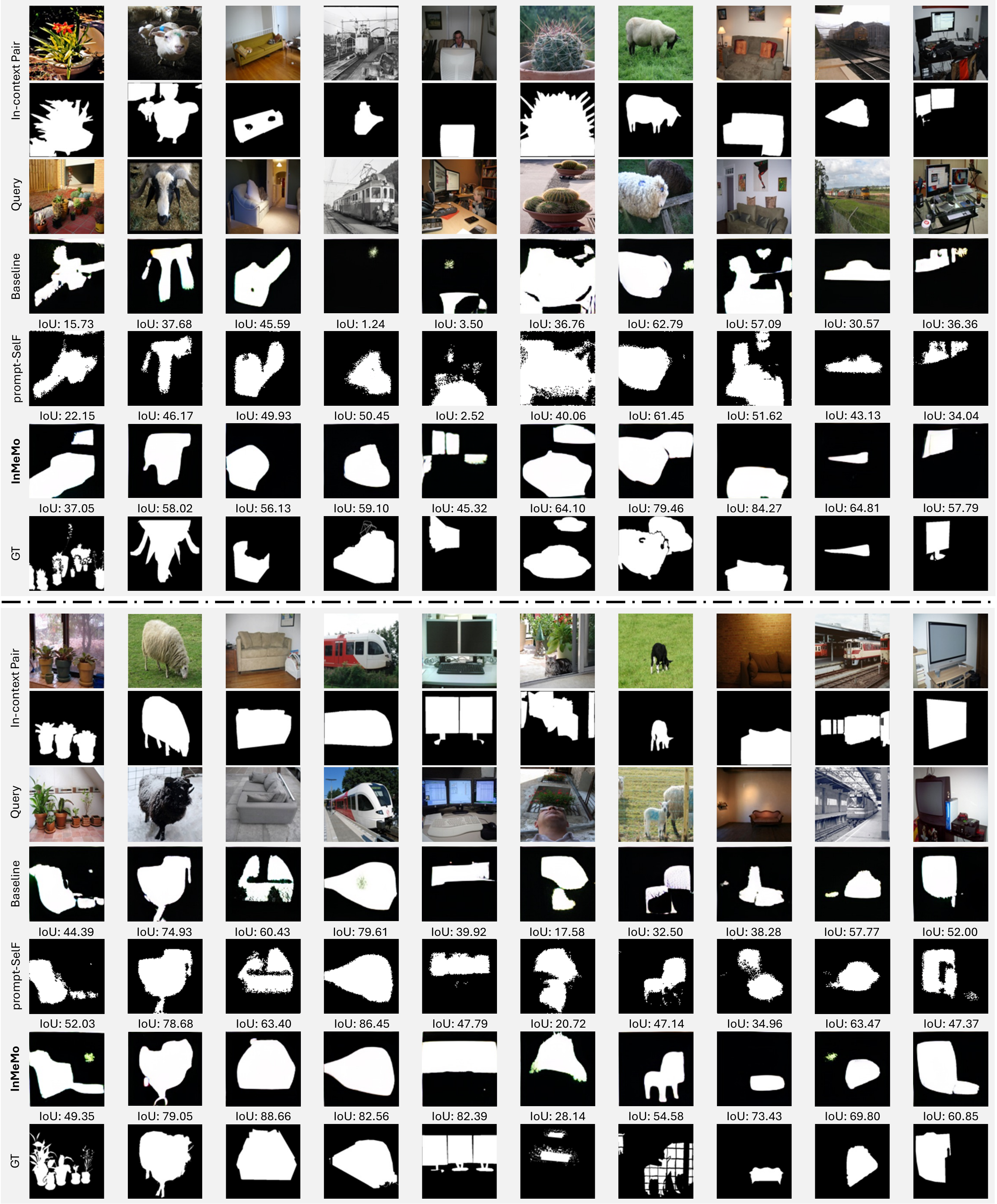}
   \caption{The visual examples in Fold-3: \textit{pottedplant}, \textit{sheep}, \textit{sofa}, \textit{train}, \textit{tvmonitor}.}
   \label{fig:fold3}
\end{figure*}

In the foreground segmentation task, we provide visual examples of the baseline, prompt-SelF \cite{promptself}, \textbf{InMeMo}, and the ground-truth label (GT). The Fold-0 (\cref{fig:fold0}), Fold-1 (\cref{fig:fold1}), Fold-2 (\cref{fig:fold2}), and Fold-3 (\cref{fig:fold3}) are represented in this section. We have arranged the format based on the different categories (columns) and demonstrated 20 examples for each fold.

\subsection{Single object detection}

\begin{figure*}[t]
  \centering
   \includegraphics[width=1\linewidth]{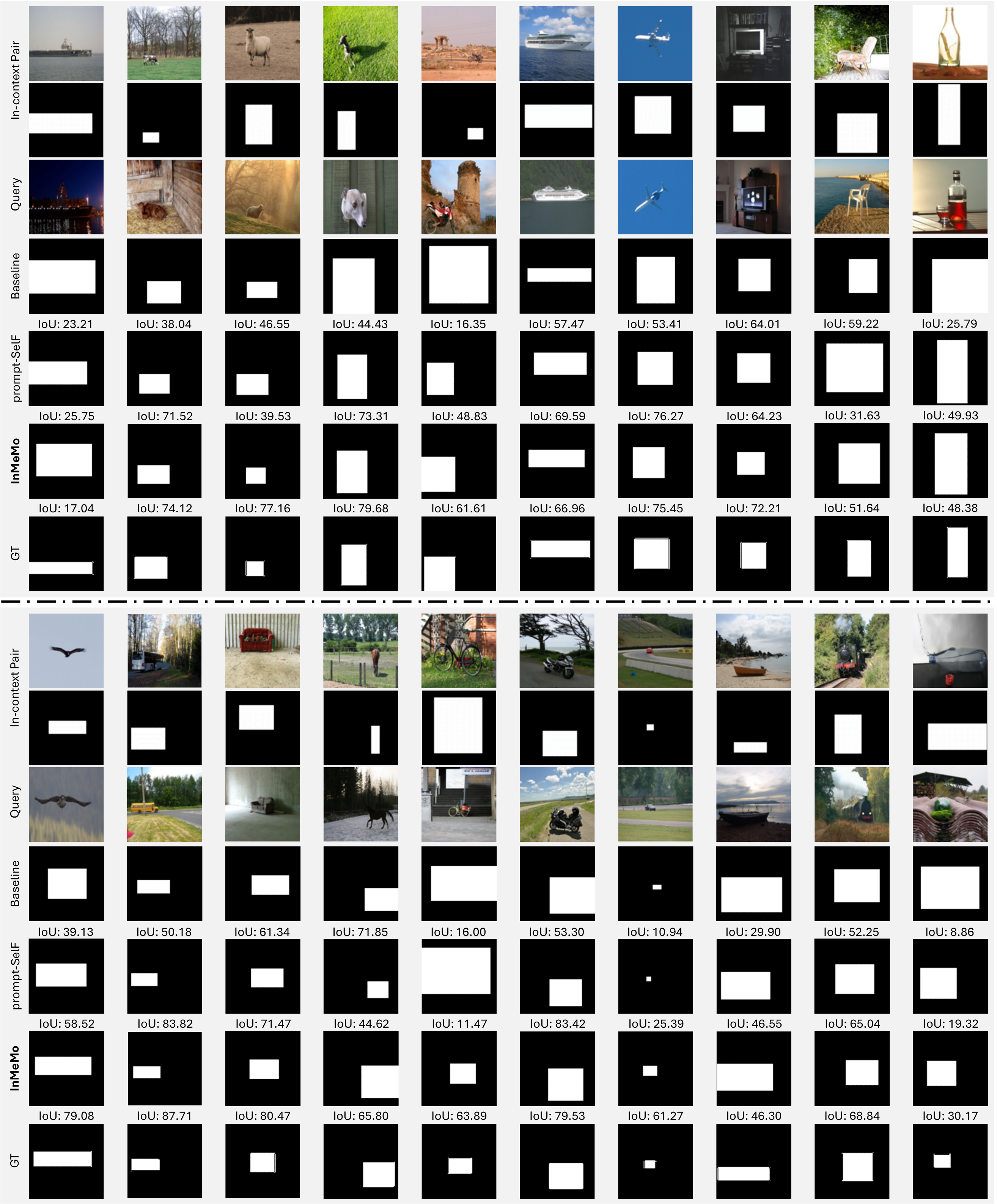}
   \caption{The visual examples in the single object detection task.}
   \label{fig:det}
\end{figure*}

We also illustrated more examples of the single object detection in \cref{fig:det}.

As mentioned in the main manuscript, InMeMo generates highly accurate details that align with ground-truth label images. Moreover, InMeMo exhibits robustness against variations in similarity, including color, size, viewpoints, and poses, between in-context and query images. Conversely, InMeMo's performance is comparable to the baseline and prompt-SelF when the similarity between the in-context and query images is high.

We have observed that InMeMo performs significantly better than previous works in single object detection tasks. This is due to our adherence to the setting proposed by \cite{maevqgan}, wherein we eliminate \textit{general} samples whose bounding box occupies more than 50\% of the entire image, to retain non-trivial samples. A more challenging scenario involves retaining only the samples in the test set where a single object occupies less than 20\% of the entire image. Therefore, such tough setting demonstrates that our InMeMo performs well on fine-grained datasets. Specifically, the in-context pairs do not adequately instruct large-scale vision models.

\end{document}